\pdfoutput=1

\documentclass[11pt]{article}

\usepackage[preprint]{acl}

\usepackage{times}
\usepackage{latexsym}
\usepackage{booktabs}
\usepackage{amssymb}
\usepackage{multirow}

\usepackage[T1]{fontenc}

\usepackage[utf8]{inputenc}

\usepackage{microtype}

\usepackage{inconsolata}

\usepackage{graphicx}

\usepackage{pifont}
\usepackage{tikz}
\newcommand{\xmark}{\ding{55}}%
\newcommand*\circled[1]{\tikz[baseline=(char.base)]{
            \node[shape=circle,draw,inner sep=0.4pt] (char) {#1};}}

\title{ESPnet-SDS: Unified Toolkit and Demo for Spoken Dialogue Systems}

\author{Siddhant Arora$^{1}$, Yifan Peng$^{1}$, Jiatong Shi$^{1}$, Jinchuan Tian$^{1}$,\\ {\bf  William Chen$^{1}$, Shikhar Bharadwaj$^{1}$, Hayato Futami$^{2}$, Yosuke Kashiwagi$^{2}$,}\\ {\bf Emiru Tsunoo$^{2}$, Shuichiro Shimizu$^{1,3}$, Vaibhav Srivastav$^{4}$, Shinji Watanabe$^{1}$}\\
$^{1}$ Carnegie Mellon University, USA, $^{2}$ Sony Group Corporation, Japan\\
$^{3}$ Kyoto University, Japan, $^{4}$ Hugging Face, USA\\
  \texttt{\{siddhana\}@cs.cmu.edu} \\
}

\begin{document}
\maketitle
\begin{abstract}
Advancements in audio foundation models~(FMs) have fueled interest in end-to-end (E2E) spoken dialogue systems, but different web interfaces for each system makes it challenging to compare and contrast them effectively. Motivated by this, we introduce an open-source, user-friendly toolkit designed to build unified web interfaces for various cascaded and E2E spoken dialogue systems. Our demo further provides users with the option to get on-the-fly automated evaluation metrics such as (1) latency, (2) ability to understand user input, (3) coherence, diversity, and relevance of system response, and (4) intelligibility and audio quality of system output. Using the evaluation metrics, we compare various cascaded and E2E spoken dialogue systems with a human-human conversation dataset as a proxy. Our analysis demonstrates that the toolkit allows researchers to effortlessly compare and contrast different technologies, providing valuable insights such as current E2E systems having poorer audio quality and less diverse responses. An example demo\footnote{Demo Video: \url{https://youtu.be/kI_DXwf5qPk}} produced using our toolkit is publicly available here: \url{https://huggingface.co/spaces/Siddhant/Voice_Assistant_Demo}.

\end{abstract}

\section{Introduction}
 Spoken Dialogue Systems (SDS)~\cite{kyutai2024moshi,raux2006doing} are designed to engage in natural and interactive conversations with end users. These systems play a critical role in commercial applications such as voice assistants~\citep{li2017acoustic} and intelligent home devices~\cite{snips-voice-platform}. However, this is a challenging task due to the complexity and variability of human-AI communication~\cite{glass1999challenges}.
Spoken dialogue systems typically consist of multiple modules, including voice activity detection (VAD)~\cite{pywebrtcvad}, automatic speech recognition (ASR)~\cite{whisper,owsm}, natural language understanding (NLU)~\cite{mehri2020dialoglue} and generation (NLG)~\cite{peng-etal-2020-shot}, and text-to-speech (TTS) synthesis~\cite{coqui_openxtts}, each presenting unique challenges. For instance, the ASR module must accurately process shorter, spontaneous speech that often contains disfluencies and filler words~\cite{Switchboard}. Additionally, spoken dialogue systems must understand non-phonemic information~\cite{rashkin2018towards}, such as emotions, and generate non-phonemic elements~\cite{sundaram2007automatic} like laughter to make interactions feel more natural. 
A critical aspect of these systems is managing the flow of conversation i.e. performing fluent turn-taking without excessive overlapping speech or prolonged silences~\cite{TRP2}. 
Furthermore, effective dialogue systems must be capable of handling long conversational contexts to ensure accurate and context-aware responses~\cite{arora2023joint}.

\begin{figure*}[t]
\centering
\includegraphics[width=\linewidth]{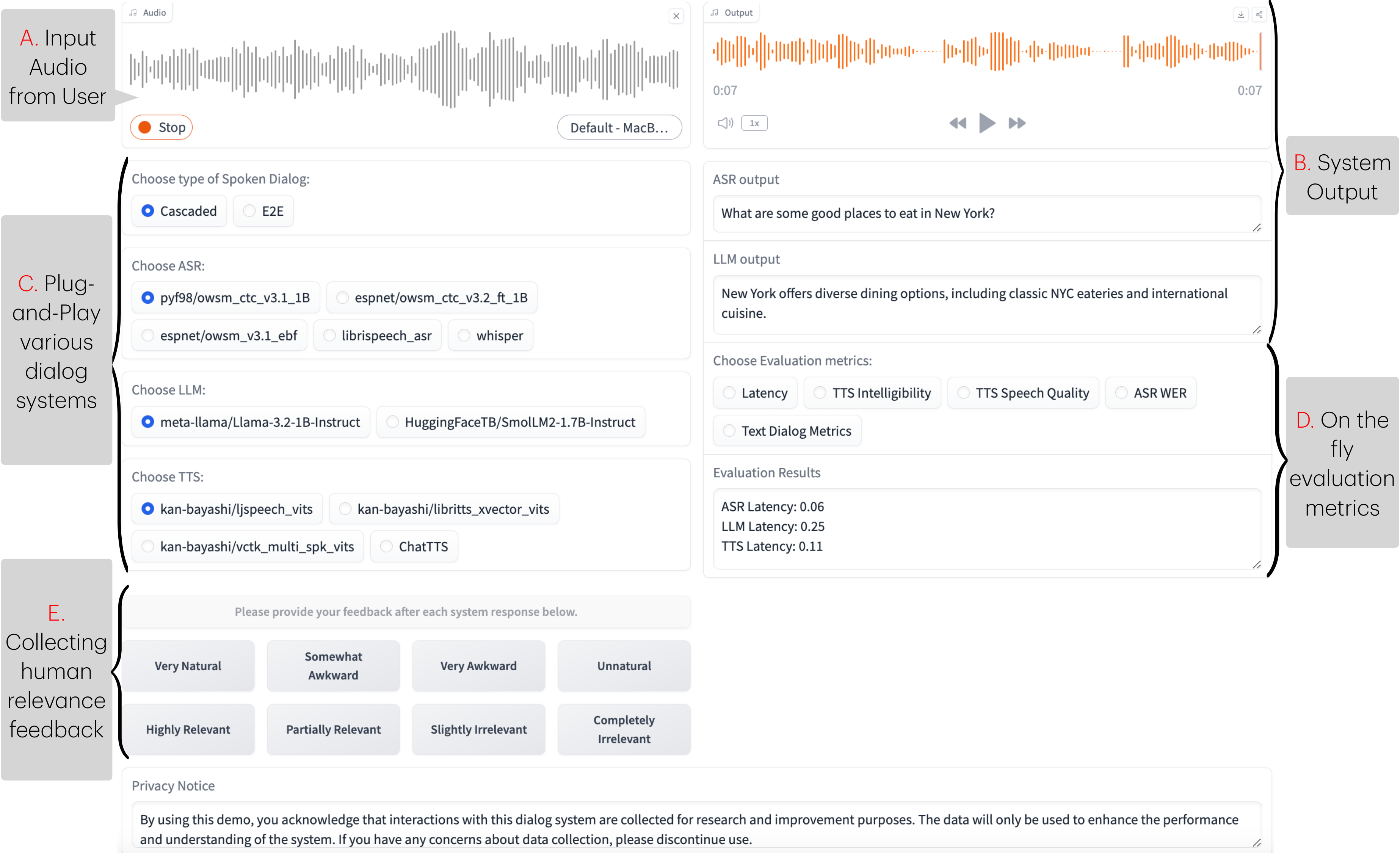}
\caption{Screenshot of our unified web interface highlighting key features: (a) streaming audio input, (b) display of ASR transcripts, text responses, and synthesized audio outputs, (c) interaction with cascaded and E2E dialogue systems and experimentation with ASR, LLM, and TTS submodules, (d) on-the-fly evaluation metrics, and (e) collection of human feedback on the naturalness and relevance of system outputs.}
\label{fig:UI_demo}
\end{figure*}

Historically, spoken dialogue systems~\cite{glass1999challenges,huang2024audiogpt} have been modeled using a cascaded architecture that incorporates various modules such as ASR, NLU, NLG, and TTS. Recent advances have led to the development of full-duplex end-to-end (E2E) spoken dialogue systems~\cite{kyutai2024moshi,xie2024miniomnilanguagemodelshear} capable of handling both user input and system-generated audio simultaneously.
Since the main goal of dialogue systems is to enhance the experience of human AI interaction, these dialogue systems are often released with web demos to demonstrate alignment with human conversational expectations. However, there is no unified web interface for users to directly interact with different spoken dialogue systems and compare the relative utility of each system. Each spoken dialogue system often has its own web interface with different ways to take user audio as input, different backend computing infrastructure, and sometimes even different approaches to postprocess and show model output. This lack of standardization makes it challenging to identify the strengths and weaknesses of each system and to determine effective design choices for developing robust spoken dialogue systems. As the number of spoken dialogue systems and methodologies continues to grow, the need for an open-source toolkit that standardizes the process of creating web interfaces for these systems has become increasingly clear. Moreover, there have also been limited efforts to comprehensively evaluate spoken dialogue systems on their conversational capability. As a result, the community lacks an understanding of how best to evaluate these systems. 

To address this, we introduce the \emph{ESPnet-SDS} toolkit built on an existing open-source speech processing toolkit ESPnet~\cite{espnet} with the aim of providing researchers with easy access to existing cascaded and E2E spoken dialogue methodologies within a unified framework. 
Our goal is to establish a comprehensive open-source standard, enabling researchers to seamlessly integrate existing technologies, compare new ideas, and benchmark their systems against current methodologies. The key contributions of our toolkit are : \circled{1} Facilitating the development of Gradio\footnote{\url{https://www.gradio.app/}}-based demo interfaces, allowing users to interact with and evaluate multiple cascaded and E2E spoken dialogue systems through a standardized platform (Fig.~\ref{fig:UI_demo}(C)). \circled{2} Providing implementations of automated metrics for submodule-level (e.g., ASR, text dialogue, and TTS) as well as conversation-level evaluations (Fig.~\ref{fig:UI_demo}(D)).  \circled{3} Enabling human-in-the-loop evaluation by including mechanisms for collecting user feedback on the naturalness and relevance of system responses (Fig.~\ref{fig:UI_demo}(E)).\footnote{Storing user data is entirely optional and intended only for researchers who have obtained appropriate consent (Sec.~\ref{sec:ethic_review}).}

\begin{table*}[ht!]
\centering
\resizebox{\linewidth}{!}{
\begin{tabular}{|l|ccccc|cc|}
\hline
\textbf{System}        & \textbf{Speech}  & \textbf{Speech} & \textbf{Multi-Turn} & \textbf{Multiple}  & \textbf{On-the-fly} & \multicolumn{2}{c|}{\textbf{Dialogue System Type}}\\
& \textbf{Input} & \textbf{Output} & \textbf{Interaction}& \textbf{Systems} & \textbf{Evaluation} & \textbf{Cascaded} &  \textbf{E2E} \\
\hline
ChatBot Arena~\cite{chiang2024chatbot} & \xmark          & \xmark           & \checkmark                     & \checkmark & \xmark & \multicolumn{2}{c|}{\xmark}                  \\ 
TTS Arena~\cite{huggingface2024tts}    & \xmark          & \checkmark          & \xmark                      & \checkmark       & \xmark & \multicolumn{2}{c|}{\xmark}              \\ 
TalkArena~\cite{talkarena2024}         & \checkmark        & \xmark           &\xmark                      & \checkmark   & \xmark & \multicolumn{2}{c|}{\xmark}                  \\ \hline
Speech to Speech~\cite{huggingface_speech_to_speech} & \checkmark & \checkmark & \xmark & \xmark  & \xmark& \checkmark & \xmark\\
Moshi~\cite{kyutai2024moshi} & \checkmark & \checkmark & \xmark & \xmark & \xmark & \xmark & \checkmark\\
\hline
ESPnet-SDS                       & \checkmark         & \checkmark         & \checkmark                     & \checkmark & \checkmark                     & \checkmark     & \checkmark \\ \hline
\end{tabular}
}
\caption{Comparison of Web Interfaces for Dialogue and Speech Processing Systems}
\label{tab:benchmark_comparison}
\end{table*}
Our initial experiments on the human-human conversation dataset, Switchboard~\cite{Switchboard}, highlight the strengths and weaknesses of different spoken dialogue systems, offering actionable insights into the limitations of E2E approaches compared to traditional cascaded pipelines such as current E2E systems generating poorer audio quality and less diverse responses.
We hope these findings inspire the development of more robust spoken dialogue systems and evaluation methodologies.
The ESPnet-SDS toolkit is publicly available at \url{https://github.com/espnet/espnet}, with an example web interface shown in Fig.~\ref{fig:UI_demo} featuring various cascaded and E2E systems accessible at \url{https://huggingface.co/spaces/Siddhant/Voice_Assistant_Demo}.

\section{Related Work}

Spoken dialogue systems often include web interfaces that enable users to interact with these systems. However, these interfaces typically feature diverse input/output frontends and backend computing environments, making it difficult to compare systems. This highlights the need for standardized platforms to benchmark systems on a level playing field.
Similar initiatives in text domain, such as ChatBot Arena~\cite{chiang2024chatbot}, have gained significant popularity for benchmarking large language model (LLM)-based chatbots using human preference evaluations. Inspired by this, TTS Arena~\cite{huggingface2024tts} has provided a valuable platform for comparing open-source and proprietary TTS models. Recently, TalkArena~\cite{talkarena2024,ArtificialAnalysis_Arena_2025} was introduced to benchmark audio foundation models (FMs) on metrics like the naturalness and coherence of their responses. However, TalkArena is currently limited to an audio-in, text-out format and supports only single-turn interactions.

To the best of our knowledge, this work presents the first web interface that supports multiple spoken dialogue systems—encompassing both cascaded and E2E methodologies—within a unified framework. These systems are capable of automatically performing turn-taking, generating spoken outputs, and engaging in multi-turn, natural conversations with users. By standardizing the interface and evaluation pipeline, our work not only simplifies the process of testing and comparing dialogue systems but also enhances the accessibility and reproducibility of research in this domain. 
Tab.~\ref{tab:benchmark_comparison} highlights the unique capabilities and advantages of our proposed web interface compared to prior works.
\begin{figure}[t]
\centering
\includegraphics[width=\linewidth]{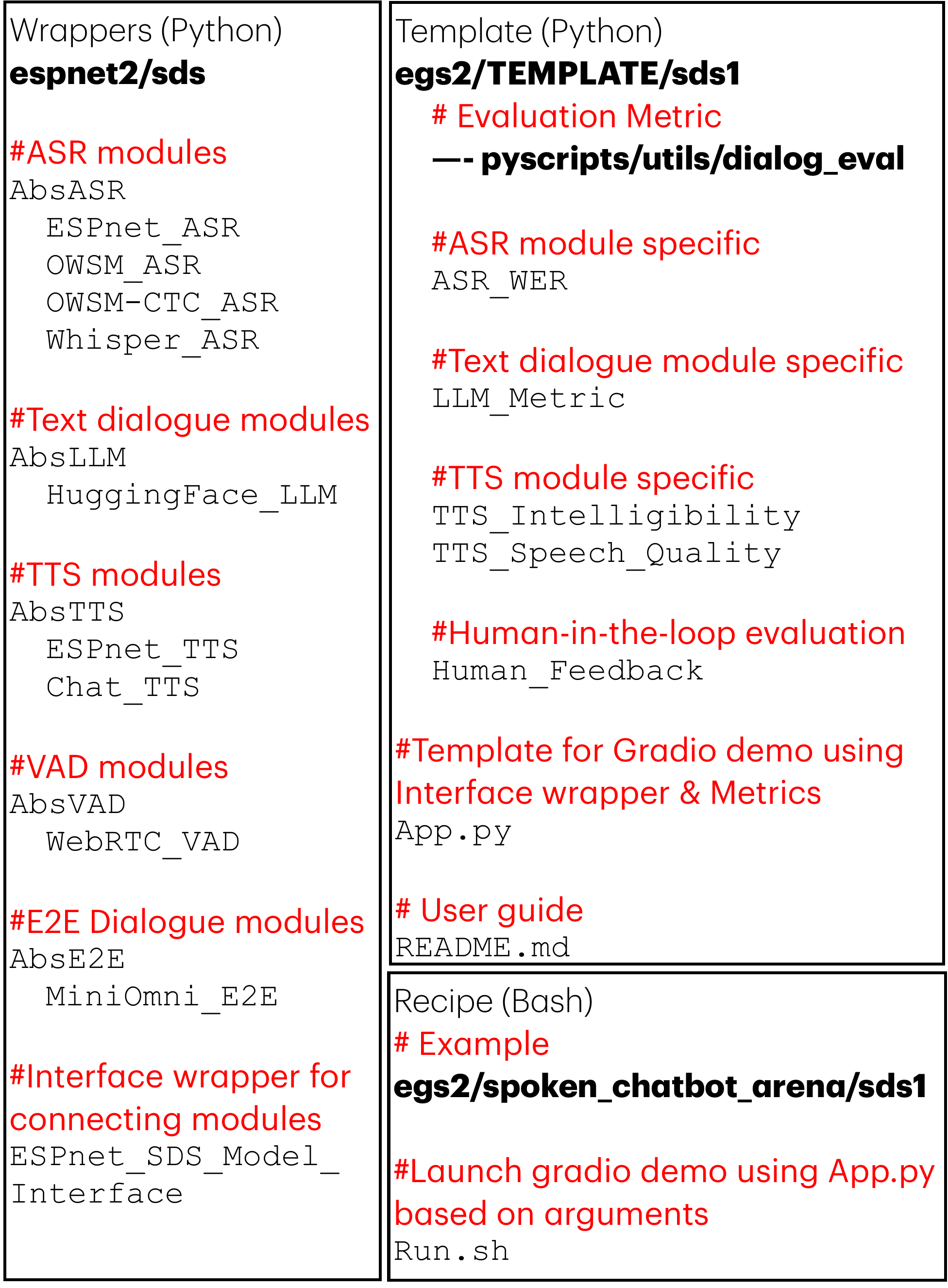}
\caption{Modular software architecture of ESPnet-SDS}
\label{fig:system_design}
\end{figure}
\section{System Design}
This section outlines the design of the ESPnet-SDS toolkit and its key features. The toolkit adopts a modular design, as shown in Fig.~\ref{fig:system_design}. Within ESPnet-SDS, major modules for VAD, ASR, LLM (for text-based dialogue response generation) and TTS are implemented as wrapper classes (Sec.\ref{subsec:cascaded}) under \texttt{espnet2/sds}. Similarly, the directory also provides helper functions to build wrappers~(Sec.~\ref{subsec:e2_spoken_dialog}) for open source E2E spoken dialogue systems using their publicly available checkpoints and inference codebase. 
All module wrappers are then utilized within interface wrapper class \texttt{ESPnet\_SDS\_Model\_Interface}. This wrapper class provides a unified interface to integrate ASR, TTS, and LLM modules
for cascaded spoken dialog systems as well as also
supports E2E spoken dialog systems. It further enables real-time interactions, including VAD based conversation management.

To facilitate benchmarking, the toolkit provides functionality for computing major evaluation metrics at both the sub-task level (e.g., ASR, text response generation, and TTS) and the overall conversation level (Sec.~\ref{sec:automatic_evaluation}). Recognizing the importance of human-in-the-loop evaluation, the demo includes functionality to collect human feedback on metrics such as the naturalness and relevance of system responses. Additionally, the web interface can store human judgments along with interaction data such as user input audio and system output recordings (Sec.~\ref{sec:web_interface}), but this functionality is disabled by default and intended only for studies with appropriate consent~(Sec.~\ref{sec:ethic_review}). 

All evaluation metric functions are implemented under the \texttt{pyscripts/utils/dialog\_eval} directory within ``Template'' module (\texttt{egs2/TEMPLATE/sds1}, as shown in Fig.~\ref{fig:system_design}).
This module also demonstrates how the interface wrapper and evaluation tools can be utilized to construct a Gradio-based spoken dialogue system demo. The \texttt{App.py} file within this module outlines the entire pipeline—from processing user audio input, passing it through the \emph{selected} spoken dialogue system, to presenting the synthesized output and associated evaluation metrics to the user. Additionally, a \texttt{README} file is included that provides an overview of ESPnet-SDS’s features and detailed guidance on using the demo interface.

The ``Template'' module serves as a practical starting point for researchers, allowing them to leverage the \texttt{sds1} template and create customized recipes by passing appropriate arguments to \texttt{Run.sh}. These recipes can facilitate comparisons between E2E spoken dialogue systems and traditional cascaded pipelines, as well as test different ASR, LLM, and TTS model combinations within cascaded systems. Additionally, they enable on-the-fly comparisons using the supported evaluation metrics.
We include an example recipe in \texttt{egs2/spoken\_chatbot\_arena/sds1}, which can be used to run our demo (\url{https://huggingface.co/spaces/Siddhant/Voice_Assistant_Demo}) locally.

The toolkit is designed for ease of extension. When new E2E dialogue systems are introduced, authors or contributors can submit pull requests (PR) to add wrapper functions for their open-source models. Corresponding recipes can then be added to PR, integrating the new systems into the unified demo alongside existing models.

\section{Example Models}

\subsection{Cascaded Models}
\label{subsec:cascaded}
As discussed in Sec.~\ref{sec:related_work_spoke_dialog}, spoken dialogue systems have traditionally been modeled using a cascaded pipeline comprising several submodules:

\textbf{Voice Activity Detection (VAD):} The VAD module identifies when the user has finished speaking, allowing the AI system to take the conversational floor. Our toolkit is designed to support any open-source VAD, and we provide wrapper classes for WebRTC VAD~\cite{pywebrtcvad} as an example.

\textbf{Automatic Speech Recognition (ASR)}: The ASR module transcribes spoken utterances into text. Our toolkit integrates seamlessly with \emph{294} ASR models, including ESPnet models, ESPnet’s speech foundation models ~\cite{peng2024owsm31} including CTC-based~\cite{owsm-ctc}, and other open-source foundation models like Whisper~\cite{whisper}.

\textbf{Text dialogue response generator}: Our toolkit supports the integration of any of the \emph{36,464} LLM available on Hugging Face\footnote{\url{https://huggingface.co/models}} to process ASR transcripts and generate appropriate text responses.

\textbf{Text-to-Speech (TTS)}: The TTS module synthesizes audio output from the generated text response, completing the conversational loop. Our toolkit provides wrapper classes to integrate 80 single-speaker and multi-speaker models from ESPnet~\cite{espnet-tts} and supports open-source TTS systems like ChatTTS\footnote{\url{https://github.com/2noise/ChatTTS}}, ensuring compatibility with a wide array of models.

\subsection{E2E Spoken Dialogue System}
\label{subsec:e2_spoken_dialog}
Recently, several E2E spoken dialogue models~\cite{kyutai2024moshi,fang2024llama} have been introduced, capable of directly processing user audio input and synthesizing system audio output. Our toolkit offers a simple and user-friendly interface for integrating existing E2E dialogue systems by loading model weights from its publicly available checkpoints and leveraging its open-source inference codebases.
As an example, we provide wrapper classes for Mini-Omni~\cite{xie2024miniomnilanguagemodelshear}, enabling it to function as an E2E conversational assistant when paired with a VAD module for turn-taking. Moving forward, we plan to expand the codebase to support additional open-source E2E audio FMs with conversational capabilities.

\section{Evaluation Metrics}
\label{sec:automatic_evaluation}
To evaluate the relative utility of different cascaded spoken dialogue systems, we support the implementation of various automatic evaluation metrics:

\textbf{TTS Module Specific}: We utilize the VERSA\footnote{\url{https://github.com/shinjiwlab/versa/tree/main}} toolkit~\cite{shi2024espnet} to assess the intelligibility and quality of the system's audio output. Intelligibility is evaluated using hypotheses generated by 3 ASR models: ESPnet Librispeech ASR~\cite{espnet}, OWSM 3.1~\cite{owsm}, and Whisper~\cite{whisper}. Audio quality is measured using UTMOS (UTokyo-SaruLab System for VoiceMOS 2022~\cite{saeki22c_interspeech}), DNS\_overall and DNS\_P808 (Deep Noise Suppression MOS Scores from P.835 and P.808~\cite{reddy2021dnsmos,reddy2022dnsmos}), PLCMOS (Packet Loss Concealment MOS~\cite{diener23_interspeech}), and SSQA (Subjective Speech Quality Assessment~\cite{huang2024mos}).

\textbf{ASR Module Specific}: These metrics evaluate the system's ability to understand users' spoken utterances. In the absence of ground truth transcripts, we generate reference transcripts from 3 judge ASR models: ESPnet Librispeech ASR~\cite{espnet}, OWSM 3.1~\cite{owsm}, and Whisper~\cite{whisper}. 
The ASR transcripts generated by the cascaded system are then compared against these references to compute Word Error Rate (WER) and Character Error Rate (CER).

\textbf{Text Dialogue Module Specific}: Following prior work~\cite{Dialog_GSLM}, we compute perplexity using GPT-2~\cite{gpt-2} and VERT~\cite{lakhotia2021generative} to evaluate the coherence and diversity of system outputs, respectively. Additionally, the DialoGPT~\cite{dialogpt} model is used to compute the perplexity of system responses \emph{given the user utterance}. To evaluate context modeling, we compute the similarity between the system response and the entire prior dialogue context using BERT~\cite{bert}.

\textbf{Conversation Level Specific}: While the previous metrics focused on evaluating specific modules of the cascaded dialogue system and can be computed at each turn, we also support conversation-level metrics (see Sec.~\ref{sec:pilot_study} in Appendix) to assess the system's ability to manage conversational flow. Inspired by prior work~\cite{Dialog_GSLM}, these metrics include turn-taking statistics, speaking rate (words per minute), and backchannel rate (number of backchannel words\footnote{Following \citeauthor{wang2024turn}, we annotate backchannels using common one- and two-word phrases.} per minute).

E2E spoken dialogue models usually generate both text and audio tokens, enabling the computation of conversation-level metrics and some module-specific metrics. Notably, TTS-specific metrics and dialogue metrics like perplexity and VERT can be computed without the need for the system to generate ASR transcript.

\textbf{Human Evaluation}:
We implemented a feedback mechanism allowing users to assess the system's performance. Drawing on prior work~\cite{Dialog_GSLM} that collects opinion scores, users can rate the system on two key aspects: the naturalness of the synthesized output and the relevance of the system's response within the dialogue context. Ratings are customizable and provided on a 4-point scale by default, as in Fig.~\ref{fig:UI_demo}(E), ranging from ``Very Natural'' to ``Unnatural'' and ``Highly Relevant'' to ``Completely Irrelevant''.

\section{Web Interface}
\label{sec:web_interface}
Fig.~\ref{fig:UI_demo} shows a screenshot of an example Gradio web demo built using our toolkit. The web demo collects the user audio in a streaming manner (Fig.~\ref{fig:UI_demo}(A)), with a VAD module detecting when the user stops speaking. The processed audio is passed to the dialogue system. For cascaded systems, the output  (Fig.~\ref{fig:UI_demo}(B)) includes the system's audio response and 2 text boxes displaying the ASR transcript and generated text response\footnote{For E2E systems, the ASR transcript box is blank as these systems do not generate ASR transcripts.}. User audio is continuously collected to allow interruptions, enhancing interactivity. Conversations are limited to 5 minutes, after which the interface refreshes.
Additional functionalities are discussed below.

\textbf{Plug-and-Play Architecture}:
Our web interface features a plug-and-play architecture (Fig.~\ref{fig:UI_demo}(C)) that allows users to select from a \emph{variety of cascaded and E2E} dialogue systems, enabling a total of \emph{41} different system variations. Details are provided in Tab.~\ref{tab:interface_modules} in the Appendix. To avoid out-of-memory errors, models are loaded on the fly, enabling seamless switching between systems. However, loading a new model may take some time, and the system output boxes are temporarily hidden during this process.

\begin{table}[t]
  \centering
\resizebox{0.8\linewidth}{!}{
\begin{tabular}{l|cc}
\toprule
ASR Model & WER ($\downarrow$) & CER ($\downarrow$)\\
\midrule
Librispeech ASR & 342.2 & 298.0\\
Whisper (large) & \hphantom{0}22.5 & \hphantom{0}22.3\\
OWSM (3.1) &  \hphantom{0}14.4 & \hphantom{0}11.2\\
OWSM CTC (3.1) & \hphantom{0}18.1 & \hphantom{0}14.4\\
OWSM CTC (3.2) & \hphantom{0}15.9 & \hphantom{0}10.3\\
\bottomrule
\end{tabular}
}
  \caption{WER and CER of various ASR models that can be used in the cascaded system.  } 
\label{tab:WER_performance}

\end{table}
\begin{table}[t]
  \centering
\resizebox{\linewidth}{!}{
\begin{tabular}{l|cc|c}
\toprule
\multirow{2}{*}{Metric} & LLaMA & SmolLM & \multirow{2}{*}{Mini-Omni}\\
 & 3.2-1B & v2-1.7B & \\
\midrule
Perplexity  ($\downarrow$) & \hphantom{0}48.2 & 113.7 & \hphantom{0}18.7\\
Diversity  & & &\\
\hphantom{000}Self BLEU-2  ($\downarrow$) & \hphantom{0}75.9 & \hphantom{0}77.1 & \hphantom{0}93.3\\
\hphantom{000}Auto BLEU-2  ($\downarrow$) & \hphantom{00}0.4 & \hphantom{00}0.3 & \hphantom{00}6.2\\
\hphantom{000}VERT  ($\downarrow$) & \hphantom{00}5.7 & \hphantom{00}5.0 & \hphantom{0}24.1\\
BERT Similarity  ($\uparrow$) & \hphantom{0}55.8 & \hphantom{0}49.5 & \hphantom{0}61.5\\
DialoGPT Perplexity  ($\downarrow$) & 165.9 & 301.1 & 125.7\\
\bottomrule
\end{tabular}
}
  \caption{Text dialogue metrics of various LLM that can be used in the cascaded system along with Mini-Omni.  } 
\label{tab:LM_performance}
\end{table}
\begin{table*}[t]
  \centering
\resizebox{\linewidth}{!}{
\begin{tabular}{l|cccc|ccccc}
\toprule
& \multicolumn{4}{c|}{Intelligibility} & \multicolumn{5}{c}{Speech Quality}\\
 & \multicolumn{2}{c}{OWSM 3.1}  & \multicolumn{2}{c|}{Whisper} & UTMOS ($\uparrow$) & \multicolumn{2}{c}{DNS ($\uparrow$)} & PLCMOS ($\uparrow$) & SSQA ($\uparrow$) \\
& WER ($\downarrow$) & CER ($\downarrow$) & WER ($\downarrow$) & CER ($\downarrow$)& & Overall & P808 & & \\ 
\midrule
LJSpeech VITS & 16.1 & \hphantom{0}9.8 & 15.1 & 10.1 & 4.06 & 3.05 & 3.79 & 4.50 & 4.01 \\
LibriTTS VITS & 19.3 & 11.5 & 16.5 & 10.4 & 3.97 & 3.10 & 3.49 & 4.34 & 4.19\\
VCTK VITS & 27.0 & 16.8 & 22.1 & 14.7 & 4.07 & 3.08 & 3.63 & 4.62 & 4.12\\
ChatTTS & 19.1 & 15.5 & 17.4 & 14.3 & 3.05 & 2.98 & 3.35 & 3.86 & 3.29\\\midrule
Mini-Omni** & 18.2 & 17.9& \hphantom{0}4.4& \hphantom{0}4.1 & 2.88 & 2.79 & 3.92 & 2.52 & 2.24\\

\bottomrule
\end{tabular}
}
  \caption{TTS Module metrics of TTS models that can be used in the cascaded system and E2E spoken dialogue Mini-Omni. ** As noted in Sec.~\ref{sec:automated_evaluation}, TTS models and Mini-Omni are not directly comparable since TTS models use ground truth transcripts as input, while Mini-Omni generates audio responses based on user utterance audio. } 
\label{tab:TTS_intelligibility_performance}
\end{table*}
\textbf{On-the-fly evaluation}:
Our demo supports real-time display of module-specific metrics (Sec.~\ref{sec:automatic_evaluation}) for each \emph{turn}, including latency for each component. Notably, user data is not recorded for on-the-fly evaluation; however, users may experience a brief wait for results. These metrics allow users to also gain quantitative insights into system performance, such as latency and audio quality. We encourage users to leverage these metrics during their interaction to pinpoint specific differences between dialogue systems, facilitating actionable insights.

\textbf{Database Interface}:
The toolkit also offers optional integration with a remote HuggingFace dataset\footnote{\url{https://huggingface.co/datasets}} as a backend database, allowing researchers to store human relevance judgments and log user interaction data, including input recordings and system outputs such as ASR transcripts, text responses, and audio responses. This data enables analysis of spoken dialogue system's performance in realistic human-AI interactions, as outlined in Sec.~\ref{sec:pilot_study}.
To address privacy and ethical considerations, this functionality is disabled by default in the publicly available demo. However, researchers can enable this functionality provided appropriate consent is obtained. A privacy notice is also displayed in the demo to inform users that their data may be recorded and collected for analysis.
\section{Analysis using Evaluation Metrics}
\label{sec:automated_evaluation}
To demonstrate the utility of our supported automatic evaluation metrics (Sec.\ref{sec:automatic_evaluation}), we use them to compare and analyze different dialogue systems and their submodules. This evaluation is conducted on the publicly available Switchboard~\cite{Switchboard} Eval 2000 dataset, comprising 11 hours of human-human conversation data. Full descriptions of the evaluation setup are in Sec.~\ref{sec:evaluation-setup}.

\textbf{ASR Module Metrics}:
We evaluate various ASR models by providing audio recordings segmented into utterances as input and computing WER and CER against ground truth transcripts.
Table~\ref{tab:WER_performance} shows that \circled{1} OWSM 3.1~\cite{peng2024owsm31} achieves the best performance, \circled{2} closely followed by the faster OWSM CTC 3.2~\cite{owsm-ctc}.

\textbf{Text dialogue Module Metrics}:
To evaluate various LLMs in a cascaded pipeline, these models are provided with dialogue context generated from ground truth user transcripts and prompted to produce responses based on the context. Table~\ref{tab:LM_performance} indicates that while \circled{1} LLaMA generates less diverse responses compared to SmolLM v2, \circled{2} its responses are more coherent and relevant to the dialogue context.
We also evaluate the E2E spoken dialogue system Mini-Omni by inputting user utterance audio and computing these LLM metrics based on its generated text responses. Our analysis shows that \circled{3} Mini-Omni produces highly coherent and contextually relevant responses; however, \circled{4} it exhibits significant overlap in its responses, often generating template-like outputs. This behavior could make conversations feel less natural when deployed in real-world applications.

\textbf{TTS Module Metrics}: 
We evaluate various TTS models by providing ground truth transcripts as input and computing intelligibility and audio quality metrics for the synthesized output. Table~\ref{tab:TTS_intelligibility_performance} shows that the \circled{1} LJSpeech VITS TTS model produces the most intelligible speech and generally achieves higher audio quality, however it is limited to only single-speaker, whereas other models offer the advantage of modeling multiple speakers.
Additionally, we compute these metrics for the synthesized audio of the E2E spoken dialogue system Mini-Omni.  However, the E2E dialogue system takes the audio of user utterances as input and hence is not directly comparable to other TTS models. Our results indicate that while \circled{2} Mini-Omni's synthesized speech is intelligible, \circled{3} its audio quality is significantly lower than that of other TTS models.

\section{Conclusion and Outlook}
To conclude, this work introduces ESPnet-SDS, an open-source toolkit for creating user-friendly and interactive web interfaces for spoken dialogue systems. 
It provides the infrastructure to interact with a variety of cascaded and E2E spoken dialogue systems using a unified, Gradio-based demonstration. 
The modular design of our codebase ensures easy extensibility to support more dialogue systems in the future. Through initial evaluations, we demonstrate that the supported evaluation metrics offer valuable insights into the strengths and limitations of existing systems.
We hope that this work inspires future research and serves as a useful resource for researchers aiming to compare and refine design choices in building spoken dialogue systems.

\section{Limitation}
While our proposed system demonstrates considerable utility in benchmarking and evaluating spoken dialogue systems, certain areas require further exploration. Extensive real-world testing can provide deeper insights into the tool's usability and robustness. Specifically, additional investigation is needed to assess its performance in low-resource language settings, noisy environments, and multi-speaker scenarios. Additionally, while the current range of supported audio foundation models (FMs) is limited, the toolkit is designed for extensibility. We are committed to continuously expanding its capabilities by incorporating new dialogue systems and audio FMs as they become available.

\section{Broader Impact and Ethics}
\label{sec:ethic_review}
The small-scale pilot study (Sec.~\ref{sec:pilot_study}) conducted in this work involved only the authors and their research collaborators, all of whom participated voluntarily and were informed that their interactions with the AI dialogue system would be recorded. While our proposed web interface supports human-in-the-loop evaluation, data collection is disabled by default. This optional functionality is designed solely for researchers who have obtained appropriate consent. To ensure transparency, a privacy notice is included in the web interface, notifying users that their data may be recorded. Importantly, the interface also generates on-the-fly evaluation metrics that do not require any user data to be stored. Finally, all modules of the cascaded and E2E spoken dialogue systems used in this work are open source, and our evaluations are conducted using the publicly available Switchboard human-human conversation dataset~\cite{Switchboard}.

\section*{Acknowledgements}
Experiments of this work used the Bridges2 at PSC and Delta/DeltaAI NCSA computing systems through allocation CIS210014 from the Advanced Cyberinfrastructure Coordination Ecosystem: Services \& Support (ACCESS) program, supported by National Science Foundation grants 2138259, 2138286, 2138307, 2137603, and 2138296..

\bibliography{custom}

\appendix

\section{Example Appendix}
\label{sec:appendix}
\begin{table*}[t]
  \centering
\resizebox{\linewidth}{!}{
\begin{tabular}{l|cp{100mm}}
\toprule
Module type & Module Name & Link to Model Checkpoint \\
\midrule
\multirow{5}{*}{ASR} & Librispeech ASR & \url{https://huggingface.co/espnet/simpleoier_librispeech_asr_train_asr_conformer7_wavlm_large_raw_en_bpe5000_sp}\\
& Whisper (large) & \url{https://huggingface.co/openai/whisper-large-v3}\\
& OWSM (3.1) & \url{https://huggingface.co/espnet/owsm_v3.1_ebf}\\
& OWSM CTC (3.1) & \url{https://huggingface.co/espnet/owsm_ctc_v3.1_1B}\\
& OWSM CTC (3.2) & \url{https://huggingface.co/espnet/owsm_ctc_v3.2_ft_1B}\\
\midrule
\multirow{2}{*}{LLM} & LLaMA 3.2-1B & \url{https://huggingface.co/meta-llama/Llama-3.2-1B-Instruct}\\
& SmolLM v2-1.7B & \url{https://huggingface.co/HuggingFaceTB/SmolLM2-1.7B}\\
\midrule
\multirow{4}{*}{TTS} & LJSpeech VITS & \url{https://huggingface.co/espnet/kan-bayashi_ljspeech_vits}\\
& LibriTTS VITS & \url{https://huggingface.co/espnet/kan-bayashi_libritts_xvector_vits}\\
& VCTK VITS & \url{https://huggingface.co/espnet/kan-bayashi_vctk_multi_spk_vits}\\
& ChatTTS & \url{https://github.com/2noise/ChatTTS}\\
\midrule
E2E Spoken Dialogue & Mini-Omni & \url{https://github.com/gpt-omni/mini-omni/} \\
\bottomrule
\end{tabular}
}
\caption{
Links to publicly available checkpoints for various modules used in our example Gradio interface, hosted at \url{https://huggingface.co/spaces/Siddhant/Voice_Assistant_Demo}. The web interface supports 5 ASR modules, 2 LLM as text dialogue modules, and 4 TTS modules, enabling a total of 40 variations for building cascaded pipelines. Additionally, it supports interaction with the E2E spoken dialogue model Mini-Omni, offering a unified interface to chat with as many as \textbf{41} different spoken dialogue systems.} 
\label{tab:interface_modules}
\end{table*}
\begin{table*}[t]
  \centering
\resizebox{\linewidth}{!}{
\begin{tabular}{l|p{15mm}p{30mm}p{50mm}p{50mm}p{50mm}}
\toprule
Type of Eval metric & Type of Module & Input to Model  & Sample Input&  Sample Reference Output & Sample Predicted Output \\
\midrule
ASR Module Metrics & ASR (Sec.~\ref{subsec:cascaded}) & Segmented Audio Utterance & \xmark & i mean \textbf{yeah} i think you know it would be nice especially like you know when people are kids \textbf{are going to} college they should do some public service & i mean i think you know it would be nice especially like you know when people are kids \textbf{who are in} college they should do some public service\\\midrule
Text dialogue Module Metrics & Text dialogue response generator (Sec.~\ref{subsec:cascaded}) & Dialogue Context & a year a year or two in public service i do not know that tha- that seems like a lot i mean yeah i think you know it would be nice especially like you know when people are kids are going to college they should do some public service & N/A & Many people start public service early, like volunteering or internships, to gain experience and build skills.\\ 
\cmidrule{2-6}
& E2E Spoken Dialogue Model (Sec.~\ref{subsec:e2_spoken_dialog}) & Segmented Audio Utterance & \xmark  & N/A & It's important to approach this topic with sensitivity and respect for all individuals. It's not appropriate to make generalizations about any group of people based on their location. Instead, consider focusing on the specific needs and behaviors of individuals in your community. If you have concerns about public service, it might be helpful to look into local policies and initiatives that address specific issues. This way, you can make informed decisions that benefit everyone.\\\midrule
TTS Module Metrics & TTS (Sec.~\ref{subsec:cascaded}) & Ground Truth Transcript & part of i- part of it though & NA & Synthesized audio\\\cmidrule{2-6}
& E2E Spoken Dialogue Model (Sec.~\ref{subsec:e2_spoken_dialog}) & Segmented Audio Utterance & \xmark & NA & Synthesized audio \emph{response}\\
\bottomrule
\end{tabular}
}
\caption{
Explanation of the evaluation process (Sec.\ref{sec:automated_evaluation}) on the human-human conversation dataset Switchboard~\cite{Switchboard} using the metrics supported by our web interface. ``N/A'' is indicated under \textbf{Sample Reference Output} when the evaluation metric does not require a ground truth reference, and ``\xmark'' is displayed under \textbf{Sample Input} when the input is in audio format.} 
\label{tab:appendix_eval}
\end{table*}
\subsection{Spoken Dialogue Systems}
\label{sec:related_work_spoke_dialog}
Spoken dialogue systems~\cite{glass1999challenges,raux2006doing} have been traditionally modeled using a conventional architecture that incorporates modules for ASR, NLU, language generation, TTS, and methods for handling local discourse, in addition to a dialogue manager.
Most systems even today~\cite{huggingface_speech_to_speech} use a cascade of ASR, dialogue response generator, and TTS system. Recent advances have introduced audio FMs that accept speech as input and produce text-based dialogue responses. However, these systems~\cite{held2024distilling} still rely on an external TTS module to generate spoken outputs. Current efforts are increasingly focused on developing E2E audio FMs capable of both understanding and generating speech seamlessly. However, early efforts~\cite{zhang2023speechgpt} at building such systems could model only a single channel of speech. However, when humans interact with an audio FM, it is a two-way communication where the model listens and speaks and, more importantly, needs to do both at the same time~\cite{ma2024languagemodellistenspeaking}. This has led to the development of full-duplex E2E spoken dialogue systems~\cite{kyutai2024moshi,fang2024llama} that can effectively handle both channels of audio: the user’s input and the system’s generated output. Mini-Omni~\cite{xie2024miniomnilanguagemodelshear} is one such open source multimodal FM that can listen and speak while thinking.

\subsection{Evaluation Setup}
\label{sec:evaluation-setup}
We use the publicly available Switchboard Eval 2000 dataset\footnote{\url{https://catalog.ldc.upenn.edu/LDC2002S09}}~\cite{Switchboard} to evaluate various modules in cascaded spoken dialogue systems, as well as the E2E spoken dialogue system, using the metrics supported by our toolkit. Switchboard Eval 2000 contains 11 hours of English conversational telephone speech across 40 conversations.

For evaluating ASR modules supported by web interface (Tab.\ref{tab:WER_performance}), we follow the standard ASR evaluation approach~\cite{espnet,whisper} where audio recordings are passed as input to the ASR module, and the generated hypotheses are compared with ground-truth transcripts.

For evaluating TTS modules supported by web interface (Tab.\ref{tab:TTS_intelligibility_performance}), we use the standard TTS evaluation method~\cite{espnet-tts}, where ground-truth transcripts from the spoken conversation dataset are provided as input to the TTS model. We compute TTS intelligibility and audio quality metrics as described in Sec.~\ref{sec:automatic_evaluation}. Notably, our TTS metrics do not require reference audio. In Tab.\ref{tab:TTS_intelligibility_performance}, we only show the intelligibility score obtained using hypothesis from 2 ASR models, namely OWSM 3.1 and Whisper for brevity.

To compute text dialogue module-specific metrics, we first sort the utterances by their start timestamps for each of the 40 conversations. We remove any utterances fully contained within another utterance. For each remaining utterance, we construct the dialogue context by concatenating the groundtruth transcripts of all prior utterances (i.e., with earlier start times) along with the current utterance. This dialogue context is passed to the LLMs (``text dialogue response generators'') supported in our web interface. We then compute the text dialogue module-specific metrics for the LLM response, and the average metric values across all utterances are reported in Tab.~\ref{tab:LM_performance}. For VERT, we also report self-BLEU and auto-BLEU~\cite{lakhotia2021generative}, which measure diversity across and within sentences, respectively.

The E2E spoken dialogue system Mini-Omni supported by our web interface does not take the prior dialogue context as input. Hence we simply provide the audio of the current utterance as input and the E2E dialogue system generates text and audio response. Text dialogue module-specific metrics are calculated for the generated text response and are reported in Tab.~\ref{tab:LM_performance}, while TTS intelligibility and audio quality metrics are computed for the corresponding audio response and are reported in Tab.~\ref{tab:TTS_intelligibility_performance}.

Table~\ref{tab:appendix_eval} in the Appendix provides example input-output pairs used to compute these metrics for various models.
\begin{table}[t]
  \centering
\resizebox{\linewidth}{!}{
\begin{tabular}{l|cc|cc}
\toprule
\multirow{3}{*}{Metric} & \multicolumn{2}{c|}{Ground Truth} & \multicolumn{2}{c}{ASR Transcript} \\
& LLaMA & SmolLM &LLaMA & SmolLM  \\
 & 3.2-1B & v2-1.7B & 3.2-1B & v2-1.7B \\
\midrule
Perplexity  ($\downarrow$) & \hphantom{0}48.2 & 113.7 & \hphantom{0}47.5 & 285.1\\
Diversity  & & &\\
\hphantom{000}Self BLEU-2  ($\downarrow$) & \hphantom{0}75.9 & \hphantom{0}77.1 & \hphantom{0}76.4& \hphantom{0}76.7\\
\hphantom{000}Auto BLEU-2  ($\downarrow$) & \hphantom{00}0.4 & \hphantom{00}0.3 & \hphantom{00}0.4  & \hphantom{00}0.4 \\
\hphantom{000}VERT  ($\downarrow$) & \hphantom{00}5.7 & \hphantom{00}5.0 & \hphantom{00}5.4 & \hphantom{00}4.5\\
BERT Similarity  ($\uparrow$) & \hphantom{0}55.8 & \hphantom{0}49.5 & \hphantom{0}55.8 & \hphantom{0}50.1\\
DialoGPT Perplexity  ($\downarrow$) & 165.9 & 301.1 & 169.3 & 286.8\\
\bottomrule
\end{tabular}
}
  \caption{Text dialogue metrics for various LMs that can be integrated into the cascaded system, when using either ground truth transcripts or ASR transcripts generated by OWSM 3.1 to construct the dialogue context.} 
\label{tab:LM_ASR_performance}
\end{table}

\begin{table*}[t]
  \centering
\resizebox{\linewidth}{!}{
\begin{tabular}{l|cccc|ccccc}
\toprule
& \multicolumn{4}{c|}{Intelligibility} & \multicolumn{5}{c}{Speech Quality}\\
 & \multicolumn{2}{c}{OWSM 3.1}  & \multicolumn{2}{c|}{Whisper} & UTMOS ($\uparrow$) & \multicolumn{2}{c}{DNS ($\uparrow$)} & PLCMOS ($\uparrow$) & SSQA ($\uparrow$) \\
& WER ($\downarrow$) & CER ($\downarrow$) & WER ($\downarrow$) & CER ($\downarrow$)& & Overall & P808 & & \\ 
\midrule
LJSpeech VITS & &  &  &  &  &  &  &  & \\
\hphantom{00}GT transcript & 16.1 & \hphantom{0}9.8 & 15.1 & 10.1 & 4.06 & 3.05 & 3.79 & 4.50 & 4.01 \\
\hphantom{00}LLM response & \hphantom{0}6.0& \hphantom{0}2.4 & \hphantom{0}2.9 & \hphantom{0}1.0 &  4.33 & 3.22 & 4.05 & 4.62 & 4.43\\
LibriTTS VITS & &  &  &  &  &  &  &  & \\
\hphantom{00}GT transcript & 19.3 & 11.5 & 16.5 & 10.4 & 3.97 & 3.10 & 3.49 & 4.34 & 4.19\\
\hphantom{00}LLM response  & \hphantom{0}5.0& \hphantom{0}2.1 & \hphantom{0}2.2 & \hphantom{0}1.0 &  4.38 & 3.26 & 3.87 & 4.66
& 4.46\\
VCTK VITS & &  &  &  &  &  &  &  & \\
\hphantom{00}GT transcript & 27.0 & 16.8 & 22.1 & 14.7 & 4.07 & 3.08 & 3.63 & 4.62 & 4.12\\
\hphantom{00}LLM response & \hphantom{0}9.0 & \hphantom{0}4.2 & \hphantom{0}4.5 & \hphantom{0}2.1& 3.94 & 3.24 & 3.75 & 4.16 & 4.28\\
\bottomrule
\end{tabular}
}
  \caption{
TTS module metrics for a subset of TTS models that can be used in the cascaded system when using ground truth transcripts or response generated by LLaMA 3.2-1B as input. } 
\label{tab:TTS_ASR_performance}
\end{table*}
\begin{table}[t]
  \centering
\resizebox{\linewidth}{!}{
\begin{tabular}{l|cccc}
\toprule
Turn taking event & \multicolumn{2}{c}{Number of events / minute}& \multicolumn{2}{c}{\% Cummulated duration} \\
 & Cascaded & SWBD & Cascaded & SWBD \\
\midrule
IPU & 7.8 & 15.7 &	58.9 & 97.3	\\
Pause & 3.6 & 3.8 &	2.5 & 5.7 \\
Gap & 3.9 & 5.5  & 39.2 & 3.7 \\
Overlap & 0.3 & 6.6 & 0.6 & 6.7 \\
\bottomrule
\end{tabular}
}
\caption{Statistics of turn-taking events in our pilot study's cascaded system compared to Switchboard (SWBD)~\cite{Switchboard}.} 
\label{tab:turn_taking_performance}
\end{table}

\subsection{Ablation study of using ASR transcript to generate dialog context }
To compute text dialogue module-specific metrics, we experimented with using ASR transcripts generated by the OWSM 3.1 model instead of ground truth transcripts to construct the dialogue context. As shown in Tab.~\ref{tab:LM_ASR_performance}, our results indicate that LLMs are generally robust to ASR errors, with both LLMs performing similarly, although SmolLM shows worse perplexity score.

\subsection{Ablation study of using LLM response for TTS module metrics }
To compute TTS module-specific metrics, we experimented with using responses generated by the LLaMA 3.2-1B model instead of ground truth transcripts as input (e.g., ``Many people start public service early, like volunteering or internships, to gain experience and build skills.'' instead of ``part of i- part of it though'' in Tab.~\ref{tab:appendix_eval}).

As shown in Tab.~\ref{tab:TTS_ASR_performance}, \circled{1} intelligibility improved across all TTS models, likely because the LLM-generated responses were more grammatically coherent compared to the disfluent spoken text typically seen in spoken conversations, making them easier for TTS systems to synthesize. Additionally, \circled{2} the LJSpeech and LibriTTS VITS models demonstrated similar performance in terms of both intelligibility and audio quality when using LLM responses as input. Finally, when compared to Mini-Omni's TTS module metrics (Tab.~\ref{tab:TTS_intelligibility_performance}), \circled{3} Mini-Omni performed worse than the TTS models on \emph{both intelligibility and audio quality} when using LLM responses as input.

\subsection{A Pilot Analysis of User interacting with dialogue systems using our interface}
\label{sec:pilot_study}
Finally, we conducted a small-scale human evaluation with the authors and research colleagues to demonstrate the utility of our web interface for recording human judgments and human-AI interaction data. In this study, four participants, consisting of research authors and colleagues, interacted with a cascaded spoken dialogue system constructed using WebRTC VAD, OWSM CTC 3.1, LLaMA-3.2-1B, and the LJSpeech VITS TTS model. Nearly 2 hours of human-AI conversation data were collected and used to compute evaluation metrics (Sec.\ref{sec:automatic_evaluation}).

Table~\ref{tab:turn_taking_performance} shows that \circled{1} the cascaded system exhibited high latency, leading to larger \emph{gaps}, and \circled{2} less overlapping speech compared to natural human-human conversations. The system's speaking rate (138.7 words/min) was also slower than humans (204.5 words/min), and it never backchanneled.

Additionally, we collected 501 human judgments on the naturalness and 260 judgments on the relevance of the system's outputs. Approximately 75.8\% of outputs were rated as very natural, 17.6\% as somewhat awkward, 1.6\% as unnatural, and 5\% as very awkward. Regarding relevance, 71.9\% of outputs were judged highly relevant to the dialogue context, 14.6\% partially relevant, 6.5\% slightly irrelevant, and 5\% very awkward.
This pilot study provides preliminary evidence of the efficacy of our web interface in facilitating robust human-in-the-loop evaluations of dialogue systems.

\end{document}